\pdfoutput=1
\documentclass[11pt]{article}

\usepackage[]{ACL2023}

\usepackage{times}
\usepackage{latexsym}

\usepackage{inconsolata}
\usepackage{tcolorbox}
\usepackage{booktabs}
\usepackage{multirow}
\usepackage{arydshln}
\usepackage{float}
\usepackage{tabularx}
\usepackage{caption,subcaption}
\usepackage{colortbl}
\usepackage{amssymb}% http://ctan.org/pkg/amssymb
\usepackage{pifont}% http://ctan.org/pkg/pifont

\usepackage{algorithm}
\usepackage{algpseudocode}
\usepackage{amsmath}

\usepackage[T1]{fontenc}
\definecolor{Gray}{gray}{0.92}
\definecolor{LightGray}{gray}{0.96}
\definecolor{LightCyan}{rgb}{0.92,0.968,0.968}
\definecolor{amber}{rgb}{1.0, 0.75, 0.0}%
\definecolor{neuesrot}{RGB}{207, 103, 102}%
\definecolor{lightblue}{RGB}{100, 181, 246}%
\definecolor{lightgreen}{RGB}{129, 199, 132}%
\usepackage[utf8]{inputenc}

\usepackage{microtype}

\newcommand{\rparagraph}[1]{\vspace{1.6mm}\noindent\textbf{#1.}}

\usepackage{inconsolata}

\title{Capturing Symmetry and Antisymmetry in Language Models \\ through Symmetry-Aware Training Objectives}

\author{Zhangdie Yuan \and Andreas Vlachos \\
        Department of Computer Science and Technology \\ University of Cambridge \\
        \texttt{zy317,av308@cam.ac.uk}}
\usepackage{tikz}
\usetikzlibrary{positioning}
\usepackage{amssymb}

\begin{document}
\maketitle

\begin{abstract}
Capturing symmetric (e.g., \textit{country borders another country}) and antisymmetric (e.g., \textit{parent\_of}) relations is crucial for a variety of applications. This paper tackles this challenge by introducing a novel Wikidata-derived natural language inference dataset\footnote{Dataset available at \url{https://huggingface.co/datasets/MoyYuan/Asymmetricity}} designed to evaluate large language models (LLMs)\footnote{This work was originally completed in December 2023.}. Our findings reveal that LLMs perform comparably to random chance on this benchmark, highlighting a gap in relational understanding. To address this, we explore encoder retraining via contrastive learning with $k$-nearest neighbors. The retrained encoder matches the performance of fine-tuned classification heads while offering additional benefits, including greater efficiency in few-shot learning and improved mitigation of catastrophic forgetting.
\end{abstract}

\section{Introduction}

Understanding symmetric and antisymmetric relations, is pivotal in various natural language processing tasks. Symmetric relations, such as \textit{synonymy}, denote bidirectional similarity, where forward and backward entailment occur simultaneously. For example, if ``Country A borders country B'', it naturally entails that ``Country B borders country A''. Conversely, antisymmetric relations like \textit{parent-child} imply a one-way association with unidirectional entailment. Accurately capturing these relational semantics about symmetry and antisymmetry is crucial, as evidenced by their central role in NLP tasks including relation extraction, natural language inference, fact-checking, and common sense reasoning \citep{riedel2013relation, storks2019recent, sap2020commonsense, guo-etal-2022-survey}. 

However, recent studies have highlighted significant challenges in the way language models handle semantics~\citep{de2021editing, lin2022does, mccoy-etal-2023-much} and symmetric and antisyemmetric relations such as synonymy in particular~\citep{alzantot2018generating}. In addressing a different aspect of LLM limitations, \citet{de2021editing} focused on the issue of outdated or incorrect facts within these models. They introduced a novel method using Hypernetworks to edit the knowledge in LLMs. This approach allows for the correction of `bugs' or adjustment of unexpected predictions, representing a significant step towards dynamic knowledge updating in LLMs and ensuring their accuracy and relevance over time. Similarly, \citet{lin-ng-2022-bert} revealed a specific limitation in BERT's handling of transitive relations. They showed that BERT’s predictions do not fully adhere to the transitivity property of the IS-A relation, where if "A is-a B" and "B is-a C", it should entail "A is-a C". This finding points to a fundamental gap in how LLMs like BERT process hierarchical relational structures, a crucial aspect of semantic understanding. Adding to these insights, \citet{mccoy-etal-2023-much} highlighted that while text generated by LLMs is often grammatically and morphologically correct, it frequently contains semantic inconsistencies, such as self-contradictions. This finding underscores another critical area where LLMs require improvement. More recently, \citet{berglund2023reversal} pointed out a critical limitation in LLMs' ability to understand bidirectional relationships. They found that LLMs trained on statements like "A is B" often fail to infer the reverse "B is A", indicating a gap in linking entities to their definitions and vice versa. This observation highlights a significant challenge in the current state of LLMs, yet it also presents an area ripe for future research, as no existing solutions were proposed in their work. 

In this paper, we demonstrate that LLMs exhibit significant limitations in capturing symmetric and antisymmetric relations, failing to outperform a random baseline on our dataset derived from Wikidata~\citep{vrandevcic2014wikidata}. To address this, we propose retraining the encoder using contrastive learning combined with k-nearest neighbors (k-NN), incorporating symmetry-aware objectives. This approach, focusing on enhancing the encoder itself rather than adding an extra component like a classification head, offers several advantages. It leads to a more versatile and reusable encoder that can be effectively applied to other tasks without the need for task-specific fine-tuning.

Our findings indicate that retraining the encoder with these symmetry-aware objectives matches the performance of fine-tuned classification heads. Moreover, this method shows improved efficiency in few-shot learning scenarios, requiring only a few training data points for each semantic relation. Additionally, our approach demonstrates better retention of previously acquired knowledge, outperforming traditional methods by up to 5.4\% in mitigating catastrophic forgetting~\citep{french1999catastrophic}. This efficiency, especially with minimal training data, underscores the effectiveness of our approach in enhancing the foundational encoding capabilities of LLMs for a broad range of NLP tasks.

\section{Symmetry-Aware training}\label{methodology}

\begin{table*}[ht!]
\centering
\resizebox{\textwidth}{!}{
\begin{tabular}{|l|l|l|l|l|}
\hline
\rowcolor[HTML]{EFEFEF} 
\textbf{Method/Property} & \textbf{\begin{tabular}[c]{@{}l@{}}Retraining with\\ Random Label Embeddings\end{tabular}} & \textbf{\begin{tabular}[c]{@{}l@{}}Retraining with k-NN\end{tabular}} & \textbf{\begin{tabular}[c]{@{}l@{}}Retraining with k-NN and\\ Learnt Distance Metric\end{tabular}} & \textbf{Standard Fine-Tuning} \\ \hline
\textbf{Encoder Functionality} & Single sentence encoding & Single sentence encoding & Pair of sentences encoding & Pair of sentences encoding \\ \hline
\textbf{Trained Parameters} & Encoder only & Encoder only & Encoder only & Encoder and head \\ \hline
\textbf{Training Objective} & Symmetry-aware training & Symmetry-aware training & Symmetry-aware training & Cross entropy \\ \hline
\textbf{Label Embeddings} & \begin{tabular}[c]{@{}l@{}}Randomly initialized,\\ static during training\end{tabular} & \begin{tabular}[c]{@{}l@{}}Implicitly represented\\ by cluster centroids\end{tabular} & \begin{tabular}[c]{@{}l@{}}Implicitly represented\\ by cluster centroids\end{tabular} & \begin{tabular}[c]{@{}l@{}}Absent (No explicit\\ label embeddings)\end{tabular} \\ \hline
\textbf{Distance Metric} & \begin{tabular}[c]{@{}l@{}}Fixed (Equation \ref{distancemetric})\end{tabular} & \begin{tabular}[c]{@{}l@{}}Fixed (Equation \ref{distancemetric})\end{tabular} & \begin{tabular}[c]{@{}l@{}}Dynamically learned\\ by the encoder\end{tabular} & \begin{tabular}[c]{@{}l@{}}Implicitly learned by\\ the classification head\end{tabular} \\ \hline
\textbf{Probing Method} & \begin{tabular}[c]{@{}l@{}}Nearest label embedding\\ selection, no k-NN\end{tabular} & \begin{tabular}[c]{@{}l@{}}k-NN for top-k closest\\ neighbors, followed by\\ majority voting\end{tabular} & \begin{tabular}[c]{@{}l@{}}k-NN for top-k closest\\ neighbors, followed by\\ majority voting\end{tabular} & \begin{tabular}[c]{@{}l@{}}Argmax on output of\\ the classification head,\\ no k-NN\end{tabular} \\ \hline
\end{tabular}
}
\caption{Comparative Analysis of Models for Capturing Symmetric and Antisymmetric Relations}
\label{tab:model_comparison}
\end{table*}

\subsection{Task Definition}
%\rparagraph{Symmetry} %The concept of symmetry can be readily illustrated through the relation "is a synonym of." For instance, if "A is a synonym of B", the reverse, "B is a synonym of A", also stands true, exemplifying a symmetric relation. Conversely, "is a parent of" is an antisymmetric relation; the statement "A is a parent of B" does not automatically infer that "B is a parent of A". 
We formally define symmetric and antisymmetric  relations following prior studies~\citep{sun2019rotate}:
\rparagraph{Definition} A \textit{relation} $\mathsf{r}$ is \textbf{\textit{symmetric}} if, for all entities $\mathsf{x}$ and $\mathsf{y}$, the relation $\mathsf{r(x,y)}$ implies $\mathsf{r(y,x)}$:
%\vspace{-3mm}
\begin{equation}
\mathsf{r_{symmetric}(x,y)} \implies \mathsf{r_{symmetric}(y,x)} \quad
%\vspace{-3mm}
\end{equation}
Conversely, a relation is \textbf{\textit{antisymmetric}} if $\mathsf{r(x,y)}$ implies the negation of $\mathsf{r(y,x)}$:
%\vspace{-3mm}
\begin{equation}
\mathsf{r_{antisymmetric}(x,y)} \implies \neg \mathsf{r_{antisymmetric}(y,x)}
%\vspace{-3mm}
\end{equation}
To test the capabilities of a large language model (LLM), we adopt a sentence-pair classification framework akin to that used in natural language inference (NLI) studies \citep{Fyodorov2000}. Our methodology entails providing the model with a sentence that explicitly states a symmetric or antisymmetric relation as the premise. The corresponding hypothesis is then formed by swapping the subject and the object in this sentence. We anticipate that a well-trained model will accurately recognize the inherent symmetry or antisymmetry in these relations, and correctly entail or contradict the hypothesis based on this understanding.

\subsection{Retraining with Random Label Embeddings}\label{models}

In our study, we utilize a base encoder as the foundational architecture for our models. Our approach focuses on retraining only the encoder, without adding any classification head or adapter layers. This means that during the retraining process, only the parameters within the encoder are modified. This strategy allows us to directly assess the encoder's ability to capture relational semantics.

The first model in our comparison (see Table \ref{tab:model_comparison}) involves retraining with random label embeddings. During training, each NLI-style sample (a pair of premise $p$ and hypothesis $h$) is fed into the model as separate sentences, and the corresponding label is provided. The objective function for this model is designed to minimize the distance for the positive label $l_+$ and maximize it with a margin for the negative label $l_-$. The objective function to minimize is given by:

\begin{equation}
d_{l_+}(p, h)^2 + \max(0, (\text{margin} - d_{l_-}(p, h)))^2
\end{equation}

A key component in our methodology is the distance metric function $d(p, h)$. Standard distance functions such as dot products and cosine similarity are not symmetry-aware (i.e., $d(p, h)$ = $d(h, p)$). To address this, we propose a symmetry-aware distance metric function derived from the RotatE model \citep{sun2019rotate}. The distance between the premise and hypothesis is calculated using by:

\begin{equation}
\label{distancemetric}
d_l(p, h) = 1 - \text{\textit{sim}}(h, (p \circ l)) = 1 - \frac{h \cdot (p \circ l)}{|h| |(p \circ l)|}
\end{equation}

This metric function, as defined in Equation \ref{distancemetric}, is instrumental in capturing both symmetric and antisymmetric properties of relations, which is central to our model comparisons.

During inference, the model computes the distance between the input sentence embedding and each label embedding, selecting the label with the closest embedding as the final classification.

\subsection{Retraining with k-NN}
The second model in our study employs k-Nearest Neighbors (k-NN) in a retraining setup. During retraining, pairs of samples (premise $p_1, p_2$ and hypothesis $h_1, h_2$) are fed into the model as separate sentences, without providing explicit labels.

The objective function aims to minimize the cosine similarity distance between label embeddings of samples with the same label (denoted as $d_+$) and maximize it with a margin for samples with different labels (denoted as $d_-$):

\begin{equation}
\label{k-NN}
d_+(l_1, l_2)^2 + \max(0, (\text{margin} - d_-(l_1, l_2)))^2
\end{equation}

Label embeddings $l_1$ and $l_2$ for each pair are calculated based on Equation~\ref{distancemetric}, as follows:

\begin{equation}
\label{labelembedding}
l_1 = h_1 \oslash p_1, \quad l_2 = h_2 \oslash p_2
\end{equation}

One advantage of the k-NN approach is its flexibility in handling a varying number of labels. Unlike fine-tuning methods, which typically require retraining with a different head for each new label, the k-NN method can adapt to new labels without extensive retraining. This flexibility is particularly beneficial for tasks like fact-checking, where claims can be ambiguous, as discussed in \citet{glockner2023ambifc}. This adaptability makes k-NN an effective tool for tasks requiring nuanced understanding and dynamic label management.

During inference for the k-NN settings, the model computes distances from the label embedding of the test sample to the label embeddings of each training sample. The final label for the test sample is determined by a majority vote among the top-k closest training samples. This approach, with both fixed and learned distance metric learning, provides a comprehensive evaluation of the encoder's ability to capture relational semantics.

\paragraph{Retraining with k-NN and Learnt Distance Metric}
In addition to the primary k-NN setting, we also explore an alternative approach where the model dynamically learns the distance metric. In this setting, pairs of samples are encoded together, and the label embedding is left for the encoder to learn. This allows the model to adaptively determine the best representation for the label embeddings based on the training data. The objective function remains the same as Equation~\ref{k-NN}. During inference, the process mirrors the primary k-NN setting, where distances are computed from the test sample's label embedding to the label embeddings of each training sample, with the final label determined by a majority vote among the top-k closest training samples. This alternative setting offers insights into the benefits of learning the distance metric in conjunction with the encoder.

% \subsubsection{Standard Fine-Tuning}
% The final model is a fine-tuning approach with a classification head on top of the language model encoder. It uses a fixed, predefined number of labels integrated into the head layer. The distance metric is learned implicitly by the classification head, and the encoder processes pairs of sentences. During inference, the model applies an argmax operation on the output of the classification head, selecting the label with the highest probability.

% Each of these models offers a unique approach to encoding relational semantics, and our comparative analysis aims to elucidate their respective strengths and limitations in the context of symmetric and antisymmetric relation understanding.

\begin{table*}[ht!]
\centering
\footnotesize
\begin{tabular}{|l|c|c|c|c|}
\hline
\rowcolor[HTML]{EFEFEF} 
\textbf{Model (Method)} & \textbf{\begin{tabular}[c]{@{}l@{}}Accuracy\\ (Lexicalized)\end{tabular}} & \textbf{\begin{tabular}[c]{@{}l@{}}Accuracy\\ (Delexicalized)\end{tabular}} & \textbf{\begin{tabular}[c]{@{}l@{}}Training\\ Samples\end{tabular}} & \textbf{\begin{tabular}[c]{@{}l@{}}Catastrophic\\ Forgetting ($\Delta$↓)\end{tabular}} \\
\hline
Random Baseline & 50\% & 50\% & - & - \\
RoBERTa-Large & 48.3\% & 49.7\% & - & - \\
RoBERTa-Large-MNLI & 51.2\% & 56.7\% & - & - \\
\hline
Random Label Embeddings & 100\% & 100\% & 48 & ↓5.8\% \\
k-NN & 100\% & 100\% & 64 & ↓7.7\% \\
k-NN with Learnt Distance Metric & 100\% & 100\% & 400 & ↓21.5\% \\
Fine-Tuning & 100\% & 100\% & 336 & ↓11.2\% \\
\hline
\end{tabular}
\caption{Accuracy (Lexicalized and Delexicalized), Training Samples, and Catastrophic Forgetting of Different Models on Symmetric and Antisymmetric Relation Tasks}
\label{tab:results}
%\vspace{-3mm}
\end{table*}

\section{Experiments and Results}\label{experiments}

\subsection{Experiment Setup}
To evaluate the understanding of symmetric and antisymmetric relations by Large Language Models (LLMs), we developed a dataset specifically tailored for this purpose, using data derived from Wikidata. This dataset consists of triples that exemplify these types of relations. Each triple was formatted to fit the Natural Language Inference (NLI) framework. To test the models' ability to generalize, we created both lexicalized and delexicalized versions of these triples. For instance, in the lexicalized format, the statement `Nibong LRT Station is part of LRT Singapore' \textit{contradicts} `LRT Singapore is part of Nibong LRT Station' due to the antisymmetric nature of the relation `is part of'. In the delexicalized version, this is represented as `Q7024230 is part of Q2231347' \textit{contradicts} `Q2231347 is part of Q7024230', using Wikidata IDs for entities. Our dataset comprises a total of 400,000 examples, spanning 14 symmetric and antisymmetric relations, with 100,000 examples set aside as test data. For a detailed description of the dataset construction process, including the steps of triple retrieval, labeling, and conversion to natural language, please see Appendix \ref{appendix:dataset_construction}.

In our experimental setup, we evaluated the retrained LLM encoders using accuracy as the primary evaluation metric. Additionally, we assessed the models for few-shot learning efficiency by recording the number of training samples needed to achieve the reported accuracy. Furthermore, to evaluate the models' resilience to catastrophic forgetting, we measured the performance drop ($\Delta$↓) on the MNLI dataset after training on our lexicalized dataset. As a point of comparison, we also included standard fine-tuning methods in our evaluation. Hyperparameters for retraining are detailed in Appendix \ref{appendix:model_hyperparameters}. For probing vanilla pre-trained models, we used a 1-nearest neighbor classifier.

\subsection{Results and Analysis}
Our study, leveraging the predefined metric, focuses on retraining the encoder of LLMs to align with complex relational semantics. The results, as shown in Table \ref{tab:results}, provide empirical evidence of the effectiveness of our approach. The table reveals that pre-trained LLMs, including those fine-tuned on MNLI, barely outperform the random baseline, indicating a significant challenge in understanding symmetric and antisymmetric relations. This is evident with RoBERTa-Large-MNLI achieving only a slight improvement over the random baseline in both lexicalized and delexicalized settings.

Interestingly, all our retraining methods achieved 100\% accuracy in both lexicalized and delexicalized formats. This outcome was anticipated, given the entity-independent nature of symmetric and antisymmetric relations in Wikidata. The high accuracy across both formats underscores the models' ability to generalize beyond specific entities, capturing the underlying relational semantics.

Our methods employing a fixed, symmetry-aware distance metric required notably fewer training samples. This efficiency suggests that with a well-defined distance metric, the task becomes more straightforward for the encoder, reducing the learning burden. In contrast, the alternative k-NN setting, where the encoder learns both representations and the distance metric, necessitated significantly more training data, as reflected in the 400 samples required for similar performance.

The fine-tuning approach, while achieving high accuracy, exhibited noticeable catastrophic forgetting, as indicated by an 11.2\% drop in performance on the MNLI dataset post-training, as shown in Table \ref{tab:results}. This aligns with previous research on catastrophic forgetting in LLMs \citep{kar2022preventing}. Notably, methods with a fixed symmetry-aware distance metric showed better knowledge retention, likely due to the reduced need for extensive retraining, thus preserving more of the previously acquired knowledge. These results highlight the potential of retraining strategies with a fixed distance metric in enhancing LLMs' understanding of complex relational semantics while mitigating issues like catastrophic forgetting.

In addition to RoBERTa-Large, our experiments also probed smaller language models such as MiniLM \citep{wang2020minilm} and further trained models all-MiniLM-v2~\citep{reimers2019sentence} with both 6 and 12 layers, which was pre-trained using over one billion sentences, including MNLI. Using different models, we achieved similar results, demonstrating the scalability and generalizability of our approach across different model sizes. More results are shown in Appendix~\ref{more}.

\section{Conclusion}
Our study reveals that retraining LLM encoders, as opposed to fine-tuning classification heads, effectively enhances the understanding of symmetric and antisymmetric relations. This was demonstrated using a novel dataset derived from Wikidata. This approach matches the performance achieved by fine-tuning and improves efficiency in few-shot learning and knowledge retention.

\section*{Limitations}
Our investigations highlight significant shortcomings in the capabilities of current LLMs, particularly in differentiating symmetric from antisymmetric relations. Despite their proficiency in various linguistic tasks, these models exhibit gaps in understanding relational semantics, especially symmetry.

The primary limitation of our study is its reliance on automatically generated datasets from Wikidata. While these datasets are factually accurate, they lack syntactic diversity, limiting the models' exposure to varied linguistic structures. Consequently, even though our results show near-perfect accuracy, we believe that the challenge of capturing symmetry and antisymmetry with LLMs is far from resolved. The dataset's focus on Wikidata may also overlook nuances present in other domains, potentially affecting performance in tasks like information extraction, question-answering, and natural language inference, where nuanced understanding is crucial.

% Despite these limitations, we present this work as a preliminary study on the linguistic phenomenon of symmetry and antisymmetry. Our findings lay the groundwork for future research, aiming to inspire more comprehensive and diverse investigations into this area. We hope that this initial exploration will lead to follow-up work that further elucidates the complexities of relational semantics in natural language processing.

\section*{Ethic Statement}
As we traverse the domain of natural language processing, ethical considerations regarding data sources and their inherent biases take center stage. In our study, both datasets are constructed from Wikidata, a collaborative, open-domain knowledge base. While Wikidata has established itself as a valuable resource in the research community, it is imperative to acknowledge the potential biases that could arise from its crowdsourced nature.

Wikidata, like many other collaboratively-curated resources, can reflect the viewpoints, biases, and potential inaccuracies of its contributors. Therefore, it's conceivable that the data extracted might bear cultural, geographical, or even individual biases. Such biases could inadvertently skew model evaluations, lead to partial perspectives, or reinforce existing stereotypes.

Researchers and practitioners employing our datasets should approach them with a cognizance of these limitations. Any results derived from models trained on such data should be interpreted with an awareness of the potential influences of the underlying biases. It is essential to promote transparency and continual scrutiny in data curation and model evaluation processes. In future iterations of this work, we recommend incorporating diverse data sources to mitigate the reliance on a single, potentially biased repository.

%\section*{Acknowledgements}
% Zhangdie Yuan and Andreas Vlachos are both supported by the ERC grant AVeriTeC (GA 865958).

\bibliography{anthology,custom}
\bibliographystyle{acl_natbib}

\appendix

\section{Detailed Dataset Construction}\label{appendix:dataset_construction}

\begin{table*}[h!]
\begin{tabular}{p{0.2\textwidth}p{0.7\textwidth}}
\toprule
Property ID &                                            Templates \\
\midrule
        P40 &                             [Y] is a child of [X]. \\
      P1382 &                   [Y] partially overlaps with [X]. \\
       P279 &                              [X] is a type of [Y]. \\
      P3373 &                           [X] is a sibling of [Y]. \\
      P1560 & [X] is an equivalent name of [Y] for other gender. \\
       P131 &                             [X] is located in [Y]. \\
        P25 &                          [Y] is the mother of [X]. \\
        P22 &                          [Y] is the father of [X]. \\
       P460 &                      [X] possibly the same as [Y]. \\
      P2670 &         [X] has part(s) that are instances of [Y]. \\
      P1542 &                                    [X] led to [Y]. \\
      P1889 &                         [X] is different from [Y]. \\
       P361 &                                [X] is part of [Y]. \\
       P828 &                                 [X] caused by [Y]. \\
\bottomrule
\end{tabular}
\caption{Templates used for natural language conversion of triples where [X] and [Y] are placeholders}
\label{table:templates}
\end{table*}

\begin{table*}[ht!]
\centering
\begin{tabular}{|l|c|c|c|c|}
\hline
\rowcolor[HTML]{EFEFEF} 
\textbf{Method} & \textbf{\begin{tabular}[c]{@{}l@{}}MiniLM\\ (6 layers)\end{tabular}}  &  \textbf{\begin{tabular}[c]{@{}l@{}}MiniLM\\ (12 layers)\end{tabular}} &  \textbf{\begin{tabular}[c]{@{}l@{}}all-MiniLM\\ (6 layers)\end{tabular}} &  \textbf{\begin{tabular}[c]{@{}l@{}}all-MiniLM\\ (12 layers)\end{tabular}} \\
\hline
\multicolumn{5}{|c|}{\cellcolor[HTML]{EFEFEF}Accuracy (Lexicalized)} \\
\hline
Pretrained & 46.7\% & 49.2\% & 51.2\% & 48.3\% \\
Random Label Embeddings & 100\% & 100\% & 100\% & 100\% \\
k-NN & 99.1\% & 99.1\% & 100\% & 100\% \\
k-NN with Learnt Distance Metric & 97.6\% & 100\% & 100\% & 100\% \\
Fine-Tuning & 99.8\% & 100\% & 100\% & 100\% \\
\hline
\multicolumn{5}{|c|}{\cellcolor[HTML]{EFEFEF}Accuracy (Delexicalized)} \\
\hline
Pretrained & 49.2\% & 48.3\% & 52.3\% & 52.3\% \\
Random Label Embeddings & 100\% & 100\% & 100\% & 100\% \\
k-NN & 100\% & 100\% & 100\% & 100\% \\
k-NN with Learnt Distance Metric & 100\% & 100\% & 100\% & 100\% \\
Fine-Tuning & 100\% & 100\% & 99.8\% & 100\% \\
\hline
\end{tabular}
\caption{Accuracy of Different Models on Symmetric and Antisymmetric Relation Tasks with MiniLM and all-MiniLM Variants}
\label{tab:additional_results}
\end{table*}

\subsection{Triple Retrieval}
We began by extracting symmetric and antisymmetric relations from Wikidata, a comprehensive community-curated open knowledge base \citep{vrandevcic2014wikidata}. To focus on the most prevalent relations, we applied a popularity filter, narrowing it down to the 14 most popular symmetric and antisymmetric relations. Using these filtered relations, we then retrieved triples from the Wikidata5m dataset \citep{wang2021kepler}.

\subsection{Labeling}
For each extracted triple, we created an additional triple by swapping the subject and the object while maintaining the same relation. These pairs of triples were then aligned with the Natural Language Inference (NLI) format, where one triple serves as the premise and the other as the hypothesis, and vice versa. If the relation in the triple is symmetric, the corresponding NLI label is set to `Entailment'; conversely, if the relation is antisymmetric, the corresponding NLI label is `Contradiction'. This approach ensures the dataset accurately reflects the inherent symmetry or antisymmetry of the relations.

\subsection{Natural Language Conversion}
The labeled triples were transformed into natural language text using manually designed prompts. The dataset was further enriched by creating lexicalized and delexicalized versions for each pair of statements, enabling the evaluation of models' ability to generalize their understanding of relations beyond specific entities. For instance, with a triple from Wikidata5M like "Nibong LRT Station" related to "LRT Singapore" with "is part of":

\textit{Lexicalized version:}
\begin{itemize}
    \item Nibong LRT Station is part of LRT Singapore.
    \item LRT Singapore is part of Nibong LRT Station.
\end{itemize}

\textit{Delexicalized version:}
\begin{itemize}
    \item Q7024230 is part of Q2231347.
    \item Q2231347 is part of Q7024230.
\end{itemize}

\section{Templates}\label{appendix:templates}
Table \ref{table:templates} presents the templates used for the natural language conversion of the triples. These templates were designed to accurately reflect the relational semantics of the properties in the dataset.

\section{Model Hyperparameters}\label{appendix:model_hyperparameters}
The hyperparameters used for the models in our study are detailed below. These values were optimized for our specific experimental setup.

\textbf{Retraining with Random Label Embeddings:}
\begin{itemize}
\setlength\itemsep{0em}
\item Learning Rate: 2e-05
\item Batch Size: 16
\item Optimizer: Adam (with default parameters)
\item Margin: 0.5
\end{itemize}

\textbf{Retraining with k-NN:}
\begin{itemize}
\setlength\itemsep{0em}
\item Learning Rate: 2e-05
\item Batch Size: 16
\item Optimizer: Adam (with default parameters)
\item Margin: 0.5
\item k: 3
\end{itemize}

\textit{Note:} The hyperparameters listed above are specific to our experimental setup and may vary based on different computational resources and model .

\section{Additional Results}\label{more}

In our comprehensive evaluation of different language models, we extended our experiments beyond RoBERTa-Large to include smaller models like MiniLM \citep{wang2020minilm} and further trained models, all-MiniLM~\citep{reimers2019sentence}. The all-MiniLM models were specifically pretrained on an extensive corpus of over one billion sentences, encompassing diverse datasets. This additional pretraining were aimed at enhancing the model's understanding of complex relational semantics as better encoders.

The results from these extended experiments are presented in Table~\ref{tab:additional_results}, showcasing the performance of MiniLM and all-MiniLM across our symmetric and antisymmetric relation tasks. These results further validate the effectiveness of our retraining approach, demonstrating its applicability and scalability across different sizes and types of models.

These findings underscore the versatility of our retraining methodology, highlighting its potential in improving the relational understanding of a wide range of language models, from large-scale models like RoBERTa to more compact models like MiniLM and all-MiniLM.

% \section{Templates}\label{Templates}
% \begin{table*}
% \begin{tabular}{p{0.2\textwidth}p{0.7\textwidth}}
% \toprule
% Property ID &                                            Templates \\
% \midrule
%         P40 &                             [Y] is a child of [X]. \\
%       P1382 &                   [Y] partially overlaps with [X]. \\
%        P279 &                              [X] is a type of [Y]. \\
%       P3373 &                           [X] is a sibling of [Y]. \\
%       P1560 & [X] is an equivalent name of [Y] for other gender. \\
%        P131 &                             [X] is located in [Y]. \\
%         P25 &                          [Y] is the mother of [X]. \\
%         P22 &                          [Y] is the father of [X]. \\
%        P460 &                      [X] possibly the same as [Y]. \\
%       P2670 &         [X] has part(s) that are instances of [Y]. \\
%       P1542 &                                    [X] led to [Y]. \\
%       P1889 &                         [X] is different from [Y]. \\
%        P361 &                                [X] is part of [Y]. \\
%        P828 &                                 [X] caused by [Y]. \\
% \bottomrule
% \end{tabular}
% \end{table*}

\end{document}